\documentclass[a4paper]{article}
\usepackage{amsmath,amstext,amsgen,amsbsy,amsopn,amsfonts,amssymb}
\usepackage{easybmat}
\usepackage{graphics}
\usepackage{hyperref}
\usepackage[pdftex]{graphicx}
\usepackage{epsfig}
\usepackage{amsthm}
\usepackage{bm}
\usepackage{algorithm}
\usepackage{algorithmic}

\usepackage{wrapfig}
\usepackage{esvect}
\usepackage{multirow}
\usepackage{array}
\usepackage{fancyvrb}
\usepackage{gensymb}

\providecommand{\keywords}[1]{\textbf{\textit{Index terms---}} #1}

\begin{document}
\title{\textbf{Learning Low-dimensional Manifolds for Scoring of Tissue Microarray Images}}

\author{
Donghui Yan$^{\dag}$, Jian Zou$^{\$}$, Zhenpeng Li$^{\ddag}$
\vspace{0.1in}\\
$^\dag$Mathematics and Data Science, UMass Dartmouth, MA, USA\vspace{0.05in}\\
$^\$$Mathematical Sciences, Worcester Polytechnic Institute, MA, USA\vspace{0.05in}\\
$^\ddag$Statistics, Dali University, Yunnan, China\\[0.05in]
}

\date{\today}
\maketitle

\abstract{Tissue microarray (TMA) images have emerged as an important high-throughput tool for cancer study 
and the validation of biomarkers. Efforts have been dedicated to further improve the accuracy of TACOMA, a cutting-edge 
automatic scoring algorithm for TMA images. One major advance is due to {\it deepTacoma}, an algorithm that incorporates 
suitable deep representations of a group nature. Inspired by the recent advance in semi-supervised learning and deep learning, 
we propose {\it mfTacoma} to learn alternative deep representations in the context of TMA image scoring. In particular, {\it mfTacoma} 
learns the low-dimensional manifolds, a common latent structure in high dimensional data. Deep representation learning and 
manifold learning typically requires large data. By encoding deep representation of the manifolds as regularizing features, 
{\it mfTacoma} effectively leverages the manifold information that is potentially crude due to small data. Our experiments 
show that deep features by manifolds outperforms two alternatives---deep features by linear manifolds with principal component 
analysis or by leveraging the group property. 
}



\section{Introduction}
\label{section:introduction}
\begin{figure}[htp]
\centering
\begin{center}
\hspace{0cm}
\includegraphics*[scale=0.36,clip]{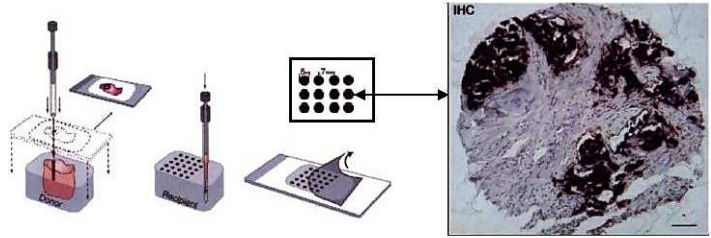}
\end{center}
\abovecaptionskip -5pt
\caption{{\it An illustration of the TMA technology (image courtesy \cite{TACOMA}). 
Small tissue cores are first extracted from tumor blocks, and stored in archives. Then thin slices of 
tissue sections are taken from the tissue core. Hundreds of tissue sections are mounted, in the form 
of an array, on a tissue slide, and stained with tumor-specific biomarkers. A TMA image is captured 
for each tissue section when viewed from a high-resolution microscope. }} 
\label{figure:tmaTech}
\end{figure}

Tissue microarray (TMA) images \cite{WanFF1987, Kononen1998, CampNR2008} have emerged as an important 
high-throughput tool for the evaluation of histology-based laboratory tests. They are used extensively in cancer 
studies \cite{Giltnane2004,CampNR2008,Hassan2008,Voduc2008}, including clinical outcome analysis \cite{Hassan2008}, 
tumor progression analysis \cite{MoussesBW2002, BeckSL2011}, the identification of diagnostic or prognostic factors 
\cite{FromontRA2005, BeckSL2011} etc. TMA images have also been used in the development and validation of 
tumor-specific biomarkers \cite{Hassan2008}. Additionally, they are used in imaging genetics \cite{CaseySBG2010,HibarKS2011} 
for the study of genetics alterations. TMA images are produced from thin slices of tissue sections cut from small 
tissue cores (less than $1$ mm in diameter) which are extracted from tumor blocks. Many slices, typically several 
hundred (possibly from different patients), are arranged as an array and mounted on a TMA slide, and then stained 
with a tumor-specific biomarker. A TMA image can be produced for each tissue section (i.e., a cell in the TMA slide) 
when viewed with a high-resolution microscope. Figure~\ref{figure:tmaTech} is an illustration of the TMA technology. 
\\
\\
Spots in a TMA image measure tumor-specific protein expression level. The readings of a TMA image are quantified by its 
staining pattern, which is typically summarized as a numerical score by the pathologist. A protein marker that is highly 
expressed in tumor cells will exhibit a qualitatively different pattern (e.g., in darker color) from otherwise. Such scores serve 
as a convenient proxy to study the tissue images. 
To liberate pathologists from intensive labors, and also to reduce the inherent variability and 
subjectivity with manual scoring \cite{Thomson2001, Hsu2002, Chung2002, 
Vrolijk2003, Giltnane2004, Berger2005,DivitoCamp2005,Walker2006, Bentzen2008,CampNR2008} 
of TMA images, a number of commercial tools and algorithms have been developed. This includes {\it ACIS, 
Ariol, TMAx} and {\it TMALab II} for IHC, and {\it AQUA} \cite{CampRimm2002} for fluorescent images. However, 
these typically require background subtraction, segmentation or landmark selection etc, and are sensitive 
to factors such as IHC staining quality, background antibody binding, hematoxylin counterstaining, and 
chromogenic reaction products used to detect antibody binding \cite{TACOMA}. The primary 
difficulty in TMA image analysis is the lack of easily-quantified criteria for scoring---{\it the staining patterns 
are not localized in position, shape or size}.
\\
\\
A major breakthrough was achieved with TACOMA \cite{TACOMA}, a scoring algorithm that is comparable to pathologists,
in terms of accuracy and repeatability, and is robust against variability in image intensity and staining 
patterns etc. The key insight underlying TACOMA is that, despite significant heterogeneity among TMA images, 
they exhibit strong statistical regularity in the form of visually observable textures or {\it staining patterns}. Such patterns 
are captured by a highly effective image statistics---the gray level co-occurrence matrix (GLCM). Inspired by the 
success of deep learning \cite{HintonSalakhutdinov2006, LeCunBengioHinton2015},
{\it deepTacoma} \cite{deepTacoma2019} incorporates deep representations to 
meet major challenges---heterogeneity and label noise---in the analysis of TMA images. The deep features 
explored by {\it deepTacoma} are of a group nature, aiming at giving more concrete information than implied by 
the labels or to borrow information from ``similar" instances, in analogy to how the cluster assumption
would help in semi-supervised learning \cite{ChapelleWeston2003, Zhu2008}.  
\\
 \\
How to further advance the state-of-the-art? Motivated by progress made with {\it deepTacoma}, 
we will further explore deep representations derived from latent structures in the data. While {\it deepTacoma} makes use of 
{\it clusters} in the data, we pursue the low-dimensional manifolds in the present work. For high dimensional data, often 
the data or part of it lie on some low-dimensional manifolds \cite{ISOMAP,LLE,DonohoGrimes2003,
HegdeBaraniuk2007,Cayton2008, BickelYan2008, JoncasMeila2013}. Effectively leveraging the manifold 
information can improve many tasks, such as dimension reduction or model fitting etc. Indeed, a recent 
work on the geometry of deep learning \cite{LeiLuoYauGu2018} attributes the success of deep learning
to the effectiveness of the deep neural networks in learning such structures in the data.
As our method is built upon TACOMA and uses manifold information, we term it {\it mfTacoma}.
\\
\\
Given the overwhelming popularity of deep learning in image recognition, it is worthwhile to remark the challenges 
in applying deep learning to TMA images \cite{deepTacoma2019}. The availability of large training sample, essential
for the success of deep learning, is severely limited for TMA images. TMA images are much harder to acquire than 
the usual natural images as they have to be taken from the human body and captured by high-end microscopes 
and imaging devices. Their labelling requires substantial expertise from pathologists. Additionally, the natural and TMA 
images are of a different nature in terms of classification. Natural images are typically formed by a visually sensible 
image hierarchy, which leads to the necessary sparsity for deep neural networks to succeed \cite{Schmidt-Hieber2017}. 
In contrast, the scoring of TMA images is not about the shape of the staining pattern, rather the ``severity and spread" 
of staining matters. A further limiting fact is that TMA images are scored by biomarkers or cancer types; there are over 
100 cancer types according to the US National Cancer Institute \cite{NCI}. 
\\
\\
Our main contributions are as follows. First, we propose an effective approach to learn the low dimensional manifolds in high
dimensional data, which allows us to advance the state-of-the-art in the scoring of TMA images. The approach is conceptually
simple and easy to implement thanks to progress in deep neural networks during the last decades. 
Second, our approach demonstrates that representing low dimensional manifolds as regularizing features is a fruitful way of 
leveraging manifold information, effectively overcoming the difficulty that shadows many manifold learning algorithms under 
small sample. Given the prevalence of low dimensional manifolds 
in high dimensional data, our approach may be potentially applicable to many problems involving high dimensional data. 
\\
\\
The remainder of this paper is organized as follows. We describe the {\it mfTacoma} algorithm in Section~\ref{section:mfTacoma}. 
In Section~\ref{section:experiments}, we present our experimental results. Section~\ref{section:conclusion} provides 
a summary of the methods and results. Given the 
\section{The {\it mfTacoma} algorithm}
\label{section:mfTacoma}
\begin{figure}[htp]
\centering
\begin{center}
\hspace{0cm}
\includegraphics*[scale=0.32,clip]{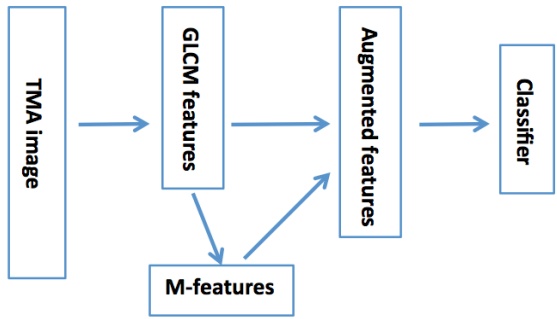}
\end{center}
\abovecaptionskip -2pt
\caption{\it Illustration of feature hierarchy in mfTacoma. } 
\label{figure:dpFeatures}
\end{figure}

In this section, we will describe the {\it mfTacoma} algorithm. The scoring systems adopted 
in practice typically use a small number of discrete values, such as $\{0,1,2,3\}$, as the score (or label) for TMA 
images. The scoring criteria are:  `$0$' indicates a definite negative (no staining of tumor cells), 
`$3$' a definitive positive (majority cancer cells show dark staining), `$2$' for positive (small portion of tumor cells 
show staining or a majority show weak staining), and `$1$' indicates weak staining in small part of tumor cells, or 
image in discardable quality \cite{Marinelli2007}. We formulate the scoring of TMA images as a classification 
problem, following (\cite{TACOMA}; \cite{deepTacoma2019}). 
\begin{figure}[htp]
\centering
\begin{center}
\hspace{0cm}
\includegraphics*[scale=0.6,clip]{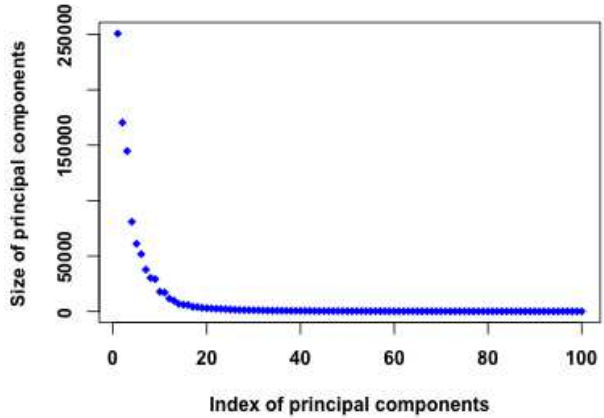}
\end{center}
\abovecaptionskip -15pt
\caption{\it Size of top 100 principal components of TMA images (in GLCM).  } 
\label{figure:pcaTMA}
\end{figure}
\\
\\
The representations we explore are derived from the low-dimensional manifold structure in the data. The 
existence of low-dimensional manifolds in high dimensional data is well established \cite{ISOMAP, 
LLE, DonohoGrimes2003, HuoNiSmith2007, HegdeBaraniuk2007,Cayton2008, BickelYan2008}. For TMA 
images, as the dimension of the ambient space (i.e., the image size) is 1504 x 1440, it is hard to ``see" the 
low-dimensional manifolds. We will instead carry out a principal component analysis (PCA) \cite{Hotelling1933, 
ShahabiYan2003, ZhaoKwok2006, Shlens2014} on the GLCM of all the TMA images to get a crude sense. 
Figure~\ref{figure:pcaTMA} shows a sharp decay of the principal components and their vanishing beyond the 50th 
component, and this clearly implies the existence of low-dimensional manifolds. Note that while PCA extracts linear 
manifolds, the actual manifolds maybe be highly nonlinear. Nevertheless, a simple PCA analysis should be fairly suggestive of their 
existence. The low-dimensional manifolds are a global property of the data and is beyond what may be revealed 
by features derived from individual data points alone. Therefore we expect such information help in the scoring of TMA images. 
One could view the information revealed by manifolds as regularization by problem structures in model fitting. Thus 
a better model (e.g., more stable or more accurate) would be expected. 
\\
\\
One technical challenge is how to extract or make use of the manifold information from the space formed 
by the TMA images. Manifold learning has been an active research area during the last two decades \cite{HuoNiSmith2007,
Cayton2008}. Many work deal with dimension reduction or data visualization, for example, Isomap \cite{ISOMAP}, 
locally linear embedding \cite{LLE}, Laplacian eigenmaps \cite{BelkinNiyogi2003}, Hessian eigenmaps 
\cite{DonohoGrimes2003}, local tangent space alignment \cite{ZhangZha2004}, diffusion maps \cite{NadlerCoifman2007}, 
metric manifold maps \cite{JoncasMeila2013}. Some of these were also used as a data-driven distance metric for 
image similarity \cite{PlessSouvenir2009}. A more fruitful line of work seems to be the use of manifolds for
regularization in model fitting. This is likely due to the difficulty in estimating the manifolds---the manifolds may be highly
nonlinear and the sample size is often disproportionally small, thus using manifolds as auxiliary information may be more 
productive. Indeed Belkin et al \cite{BelkinNiyogiSindhwani2006} successfully used the graph Laplacian as a regularizer in semi-supervised 
learning, and Osher and his colleagues \cite{OsherShiZhu2017} use the dimension of the manifold as a regularizer in image 
denoising. 
\\
\\
We follow a similar line as \cite{BelkinNiyogiSindhwani2006, OsherShiZhu2017} but with representations derived from 
the low-dimensional manifolds as {\it regularizing features} to be appended to the input. This is particularly easy
to implement, without having to solve a complicated optimization problem, and is fairly general. The effectiveness of
using regularizing features has been demonstrated in \cite{deepTacoma2019}. Thanks to the availability and easy implementation 
of autoencoder \cite{GoodfellowBengioC2016}, we will use it to extract the low-dimensional manifold representation corresponding 
to the TMA images. We term such deep representations as M-features.
Figure~\ref{figure:dpFeatures} is an illustration of the feature hierarchy in {\it mfTacoma}.
\\
\\
The {\it mfTacoma} algorithm is fairly simple to describe. First, all TMA images are converted to 
their GLCM representations. Then the GLCMs are input to the autoencoder to extract the deep features, 
to be concatenated with the GLCMs. The manifold-augmented features and their respective scores are fed to 
a training algorithm. The trained classifier will be applied to get scores for TMA images in the test set. 
To give an algorithmic description to {\it mfTacoma}, assume there are $n$ training instances, $m$ test instances, 
and $N=n+m$. Denote the training sample by $(I_1,Y_1),...,(I_n,Y_n)$ where $I_i$'s are images and $Y_i$'s are 
scores (thus $Y_i \in \{0,1,2,3\}$). Let $I_{n+1}, ..., I_{n+m}$ be new TMA images that one wish to 
score (i.e., the test set has a size of $m$). {\it mfTacoma} is described as 
Algorithm~\ref{algorithm:mfTacoma}.
\\
\begin{algorithm}
\caption{The {\it mfTacoma} algorithm}
\label{algorithm:mfTacoma}
\begin{algorithmic}[1]
\STATE {for $i=1$ to $N$} 
\STATE Compute GLCM of image $I_i$; 
\STATE Denote the resulting GLCM by $X_i$; 
\STATE endfor 
\STATE Find manifold representation $\cup_{i=1}^N \{Z_i\}$ for $\cup_{i=1}^N \{X_i\}$ with an autoencoder; 
\STATE Concatenate $X_i$ and $Z_i$ and get $X_i^M$, $i=1,...,N$;
\STATE Feed $\cup_{i=1}^n \{(X_i^M,Y_i)\}$ to RF to obtain a classification rule $\hat{f}$;
\STATE Apply $\hat{f}$ to $X_i^M$ to obtain scores for images $I_i$ for $i=n+1,...,n+m$.
\end{algorithmic}
\end{algorithm}
\\
\\
For the rest of this section, we will briefly describe the GLCM and autoencoder.

\subsection{The gray level co-occurrence matrix}
\begin{figure}[ht]
\centering
\begin{center}
\hspace{0cm}
\includegraphics*[scale=0.34,clip]{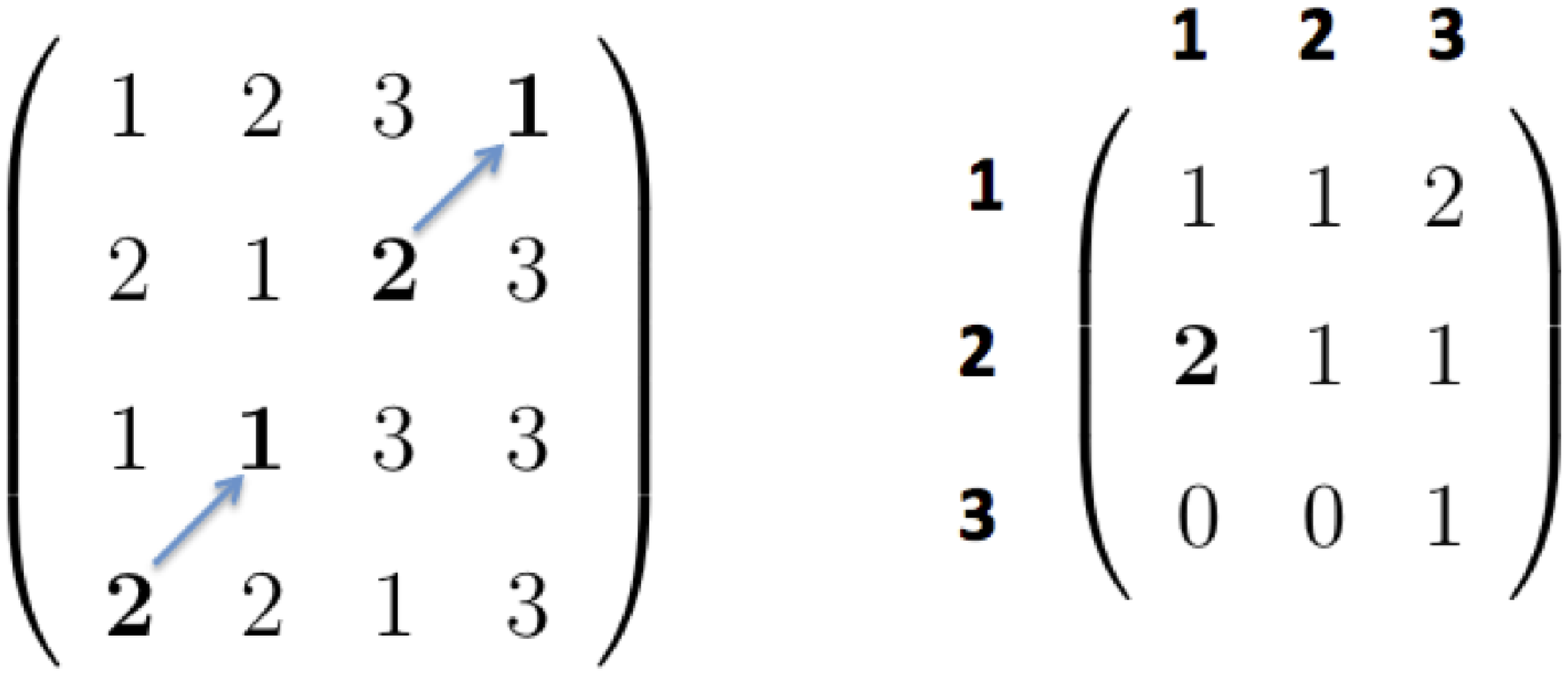}
\end{center}
\caption{{\bf Illustration of GLCM via a toy image} (image taken from \cite{deepTacoma2019}). 
{\it In the $4 \times 4$ toy image, there are three gray lavels, $\{1,2,3\}$; the resulting GLCM for spatial relationship $(\nearrow,1)$
is a $3 \times 3$ matrix.}  }  \label{figure:glcms}
\end{figure}

The GLCM is one of the most widely used image statistics for textured images, such as satellite images and 
high-resolution tissue images. It can be crudely viewed as a ``spatial histogram" of neighboring pixels in an image. 
It has been successfully applied in a variety of applications
\cite{haralick1979, GongMarceau1992,Lloyd2004,TACOMA, deepTacoma2019}. 
We follow notations used in \cite{TACOMA, YanBickelGong2018}.
\\
\\
The GLCM is defined with respect to a particular spatial relationship of two neighboring pixels. The spatial
relationship entails two aspects---a spatial direction, in set $\{\nearrow,~\searrow,~ \nwarrow, ~\swarrow, ~\downarrow,
~\uparrow, ~\rightarrow, ~\leftarrow\}$, and the distance between the pair of pixels along the direction. 
For a given spatial relationship, the GLCM for an image is defined as a $N_g \times N_g$ matrix with its $(a,b)$-entry 
being the number of times two pixels with gray values $a, b \in \{1,2,...,N_g\}$ are spatial neighbors; here $N_g$ is the 
number of gray levels or quantization levels in the image.
Note that, for each spatial relationship, one can define a GLCM thus one image can correspond to multiple GLCMs. 
The definition of GLCM is illustrated in Figure~\ref{figure:glcms} with a toy image 
(taken from \cite{deepTacoma2019}). For a balance of computational efficiency and discriminative power, we take 
$N_g=51$ and uniform quantization \cite{GrayNeuhoff1998} is applied over the $256$ gray levels.
\subsection{Autoencoder with deep networks}
\label{section:autoEncoder}
\begin{figure}[h]
\centering
\begin{center}
\hspace{0cm}
\includegraphics[scale=0.28,clip]{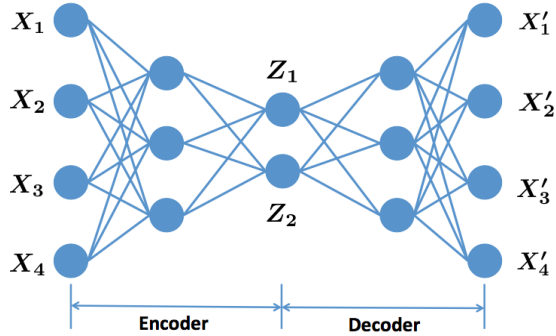}
\end{center}
\abovecaptionskip=-5pt
\caption{\it Illustration of deep neural network and autoencoder. $X_1,...,X_4$ are components of the inputs and $Z_1, Z_2$ 
are components of the codewords.} 
\label{figure:aeDiag}
\end{figure}

An autoencoder is a special type of deep neural network with the inputs and outputs being the same. A neural 
network is a layered network that tries to emulate the network of connected biological neurons in the human brain; 
it can be used for tasks such as classification 
or regression. The first layer of a neural network accepts the input signals, and the last layer for outputs. 
Here each node (unit) in the input or output layer corresponding to one component of the inputs or outputs 
when these are treated as a vector; note for simplicity here we omit the nodes corresponding to the bias terms. 
Each node in the intermediate layers (called {\it hidden layers}) takes inputs from all the connecting nodes 
in the previous layer, then performs a nonlinear activation operation and outputs the resulting signals to those connecting 
nodes in the next layer. As the data flows from one layer to the next, a weight is applied at all the connecting 
links. An illustration of the deep neural network is given in Figure~\ref{figure:aeDiag}. In the following, we will 
formally describe the details. 
\\
\\  
Assume the input data is represented by a matrix $\bm{X}_{N \times p}$, where $N$ is the number 
of data instances and $p$ is the data dimension or the number of nodes in the input layer. Let the 
weights to the links connecting the (i-1)-th and i-th hidden layer be denoted by $\bm{W}^{(i)}$, with 
its dimension determined by the number of nodes involved. Let the output signal from the i-th hidden 
layer be denoted by $\bm{Z}^{(i)}$. Let $\sigma$ be the nonlinear activation function (assume all hidden 
layers have the same activation function for simplicity). Assume there are $K$ hidden layers. Then the 
output signals at the first hidden layer are given by
\begin{eqnarray*}
&& \bm{Z}^{(1)} = \sigma \left (\bm{W}^{(1)} \bm{X} + \bm{b}^{(1)} \right).
\end{eqnarray*}
With the convention $\bm{Z}^{(0)}=\bm{X}$, the output signals at the i-th hidden layer can 
be expressed as
\begin{eqnarray*}
&& \bm{Z}^{(i)} = \sigma \left (\bm{W}^{(i)} \bm{Z}^{(i-1)} + \bm{b}^{(i)} \right), i=1,2,...K.
\end{eqnarray*}
Often the training of the neural network is formulated as solving
\begin{equation}
\arg\min_{\bm{W}, \bm{b}}J(\bm{W}, \bm{b}, \bm{X}, \bm{Y}) =  l\left(\bm{Z}^{(K)},\bm{Y}\right) - \frac{\lambda}{2} \cdot ||\bm{W}||,
\label{eq:nnFormulation}
\end{equation}
where $l(.)$ is a loss function (e.g., squared error for regression or cross entropy for classification), and 
$||.||$ is a norm such as the $L_2$ norm. The solution to \eqref{eq:nnFormulation} is often obtained by 
the {\it back-propagation} algorithm. $\bm{Z}^{(K)}$ is obtained from $\bm{X}$ by the composition of a series 
of functions 
\begin{equation*}
\sigma \circ \bm{W}^{(K)}\cdots \sigma \circ \bm{W}^{(2)}\sigma \circ \bm{W}^{(1)},
\end{equation*}
thus it can be written as $\bm{Z}^{(K)}=f_{\bm{W},\bm{b}}(\bm{X})$ for some function $f_{\bm{W},\bm{b}}$, 
which can be viewed as a smoothing function (or transformation) of the data. When the 
output $\bm{Y}$ is the same as the input $\bm{X}$, the neural network is called an autoencoder. The deep 
network in Figure~\ref{figure:aeDiag} becomes an autoencoder when all $X_i'=X_i$; $Z_1,Z_2$ are components 
of the codewords, that is, $(X_1,...,X_4) \rightarrow (Z_1,Z_2)$.
\\
\\
Although not necessary but typically an autoencoder can be divided as the encoder part and the decoder 
part which are mirror-symmetric (i.e., corresponding layers have the same number of units or connecting 
weights) w.r.t. the layer in the center. If some hidden layer has less units than that of the inputs, that means 
the data $\bm{X}$, at certain stage of its transformation, has a dimension less than the original dimension 
thus achieving a dimension reduction effect. For example, in Figure~\ref{figure:aeDiag}, the original data dimension
is 4 and the transformed data has a dimension of 2. If one is willing to assume that the activation function is 
smooth, then the data can be viewed as lying on a low dimensional manifold. The fitting of the neural network 
can thus be viewed as a way of learning the low-dimensional manifold. As the activation is nonlinear, the 
resulting manifold is also nonlinear. As individual hidden layers in a neural network can be viewed as extracting 
features of the original data, we will use such features as representation of the low dimensional manifold. 
These features are called {\it M-features}, to be appended to the existing GLCM features in the scoring of TMA 
images.
\\
\\
Note that we state in Section~\ref{section:introduction} that for TMA images, the training sample size
is often far less than that required by a typical deep neural network. However, we could still use the 
autoencoder, for two reasons. First the deep representation we will extract with an
autoencoder is to be used for regularization, thus it would be sufficient as long as the representation
captures main features of the manifolds. Second, we can control the size of the deep
network according to the training sample size; indeed we will be using the simplest autoencoder, that is, 
with only one hidden layer in this work. The algorithm for manifolds extraction is simply to extract
information of the trained hidden layer, when using some deep neural network package (The {\it deepnet} 
package is used in this work). Let $nHidden$ be the dimension of the low-dimensional 
manifold. An algorithmic description is given as Algorithm~\ref{algorithm:aeManifold}.   
\begin{algorithm}
\caption{$mfLearner(\bm{X}, nHidden)$}
\label{algorithm:aeManifold}
\begin{algorithmic}[1]
\STATE $dnn \gets  dbn.dnn.train(\bm{X}, \bm{X}, nHidden)$; 
\STATE Extract from $dnn$ a low-dimensional representation $\cup_{i=1}^N \{Z_i\}$;
\STATE Return($\cup_{i=1}^N \{Z_i\}$);
\end{algorithmic}
\end{algorithm}
\section{Experiments}
\label{section:experiments}
We conduct experiments on TMA images, the data at which our methods are primarily targeting. The TMA images 
are taken from the Stanford Tissue Microarray Database (\texttt{http://tma.stanford.edu/}, see \cite{Marinelli2007}). 
TMAs corresponding to the biomarker, estrogen receptor (ER), for breast cancer tissues are used since ER is a 
known well-studied biomarker. There are a total of $695$ such TMA images in the database, and each image is 
scored at four levels (i.e., label), from $\{0,1,2,3\}$. 
\\
\\
The GLCM corresponding 
to $(\nearrow,3)$ is used, which, according to \cite{TACOMA}, is the spatial relationship that leads to the greatest 
discriminating power for ER/breast cancer. The pathological interpretation is that, the staining pattern is approximately 
rotationally invariant (thus the choice of direction is no longer important) and `3' 
is related to the size of the staining pattern for ER/breast cancer. 
\\
\\
For autoencoder, we use the R package {\it deepnet}. Random Forests (RF) \cite{RF} is chosen as the classifier due 
to its superior performance compared to popular methods 
such as support vector machines (SVM) \cite{CortesVapnik1995} and boosting \cite{AdaBoost}, according to large 
scale simulation studies \cite{caruanaKY2008}. This is also true for the scoring (\cite{TACOMA}; \cite{deepTacoma2019}) 
and the segmentation of TMA images \cite{HolmesKapelner2009}. This is likely due to the high dimensionality (2601 
when using GLCM) and the remarkable feature selection as well as noise-resistance ability of RF, while SVM and 
boosting methods are typically prune to those. 
\\
\\
For simplicity, the test set error rate is used as our performance metric. In all experiments, a random selection 
of half of the data are used for training and the rest for test, and results are averaged over $100$ runs. 
We conduct two types of experiments. One is to use linear manifolds extracted by PCA. The other is to use
nonlinear manifolds extracted by autoencoder. 
\subsection{Experiments with manifolds by PCA}
\label{section:experimentsPCA}
\begin{figure}[htp]
\centering
\begin{center}
\hspace{0cm}
\includegraphics*[scale=0.42,clip]{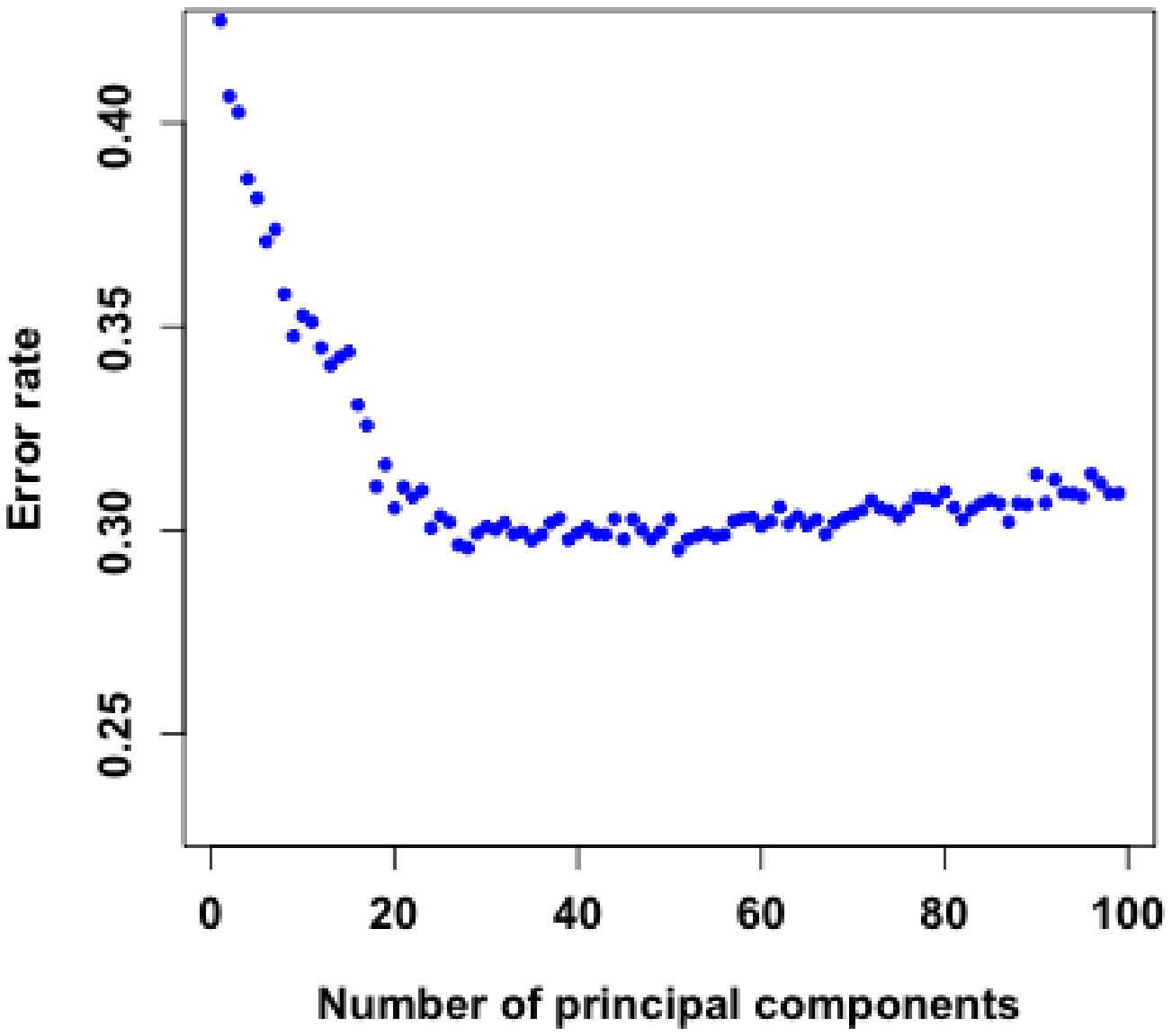}
\includegraphics*[scale=0.42,clip]{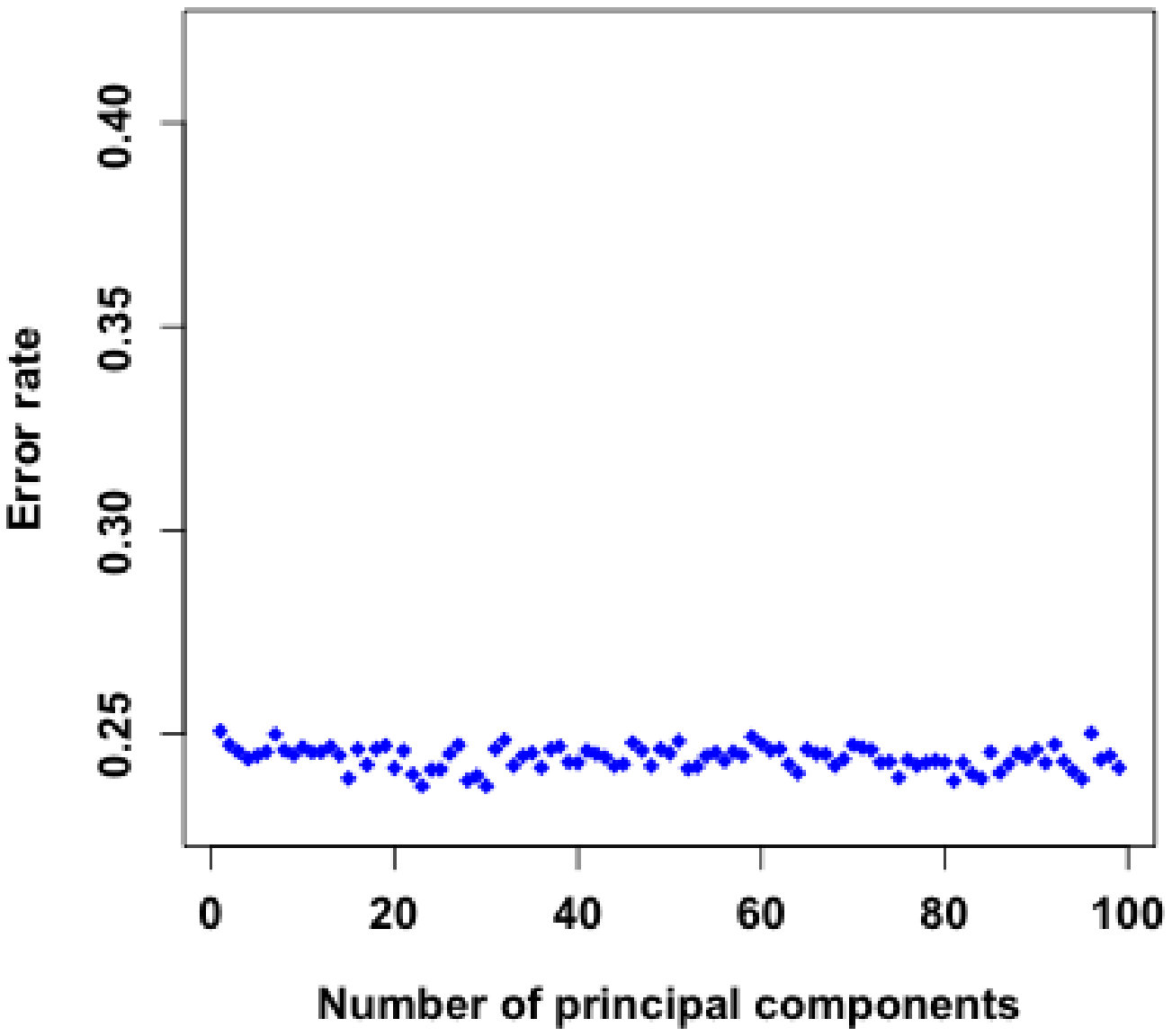}
\end{center}
\abovecaptionskip -5pt
\caption{{\it Test set error rate when including more leading principal components. Left: using only principal 
components. Right: using both GLCM features and principal components. }} 
\label{figure:pcaAccu}
\end{figure}

We mention in Section~\ref{section:mfTacoma} that one can extract linear manifolds with PCA. How effective are those 
linear manifolds in the scoring of TMA images? We distinguish between two cases. One is to use the leading 
principal components as sole features to be input the classifier, the other is to use the leading principal components 
as regularizing features appended to the GLCMs. The results are shown in Figure~\ref{figure:pcaAccu} 
where we produce the test set error rates when the number of leading principal components increases up to 100. 
It can be seen that the PCA features as regularizing features clearly outperform that using those as sole features
with a performance gap close to about 6\%.
\subsection{Experiments with manifolds by autoencoder}
\label{section:experimentsAE}
For nonlinear manifolds with autoencoder, we conduct experiments on three different cases: 1) Use all the original image pixels as features 
along with M-features extracted from the space of the original images; 2) GLCM features along with M-features extracted 
from the space of the original images; 3) GLCM features along with M-features extracted from the space of GLCMs. We also
tried using M-features extracted from the space of original images alone and that of GLCMs alone, but none gave satisfactory results. 
Table~\ref{table:accuracyTMA} shows results obtained in the three cases where the dimension of the low-dimensional 
manifold varies.
\begin{table}[h]
\begin{center}
\begin{tabular}{|c|c|c|}
    \hline
\bf{Features}                    		    &\bf{Dim. of manifold}               &\bf{Error rate}\\
\hline \hline
GLCM   &---    													                   &$24.79\%$\\
\hline
Image + AE image features  				&100                     &29.89\%\\
						   				&50					&29.88\%\\
						   				&25				 	&29.90\%\\	
\hline
GLCM + AE image features         &400		                &$24.51\%$\\
								         &300		                &$24.64\%$\\
         &200		                &$24.03\%$\\
         &100		                &$24.63\%$\\
        &50		                &$24.42\%$\\
         &25		                &$24.36\%$\\
         &10		                &$24.59\%$\\
\hline
GLCM + AE GLCM features         &200		                &$24.72\%$\\
								      	&100		                &$24.66\%$\\
								      	&50		                &$24.59\%$\\
								      	&40		                &$23.84\%$\\
								      	&30		                &$23.23\%$\\
								      	&25		                &$\bf{22.85}\%$\\
								      	&20		                &$23.08\%$\\
								      	&10		                &$23.75\%$\\
\hline
\end{tabular}
\end{center}
\caption{\it{Error rate in scoring TMA images when using different deep features under different 
manifold dimensions. Note that the first row corresponds to results obtained by RF on the GLCM 
features alone (i.e., without deep features). Here `Image' indicates when the original image pixels 
are used as features, `AE image features' indicates manifold features extracted from the space of 
the original TMA images, `AE GLCM features' indicates manifold features extracted from the space 
of the GLCM of the TMA images.  
}} \label{table:accuracyTMA}
\end{table}
\\
\\
Using original images 
pixels as features along with manifolds features extracted from the space of original images does not lead 
to satisfactory results, while using GLCMs with manifolds features from the space of original images improves 
marginally over using GLCMs alone. The best result is achieved at an error rate of 22.85\% when using GLCM 
features along with M-features extracted from the space of GLCMs with the dimension of the manifold being 25. 
Compared to an error rate of 23.28\% achieved by {\it deepTacoma} \cite{deepTacoma2019} and 24.79\% without 
using any deep features by {\it TACOMA} \cite{TACOMA}, the improvement is substantial given that the performance 
by {\it TACOMA} \cite{TACOMA} already rivals pathologists, and that progress in this area is typically incremental 
in nature.   

\section{Conclusions}
\label{section:conclusion}
Inspired by the recent success of semi-supervised learning and deep learning, we explore deep representation
learning under small data, in the context of TMA image scoring. In particular, we propose the {\it mfTacoma} algorithm 
to extract the low-dimensional manifolds in high dimensional data. Under {\it mfTacoma}, deep representations 
about the manifolds, possibly crude due to small data, are conveniently used as regularizing features to be appended 
to the original data features. This turns out to be a simple and effective way of making use of the low dimensional manifold 
information. Our experiments show that {\it mfTacoma} outperforms linear manifold features extracted by PCA or deep 
features of a group nature. We consider this a notable improvement over TACOMA and {\it deepTacoma} given that those 
already rival trained pathologists in the scoring of TMA images and progress in this area is typically incremental in nature. 
\\
\\
Given the prevalence of low dimensional manifolds in high dimensional data, we expect that deep 
features derived from low dimensional manifolds would help in many applications. Our approach of leveraging the manifold
information as regularizing features will be useful in small data setting, where one may incorporate feature weights by 
accounting for the sample size or data quality.  

\end{document}